\documentclass[10pt,twocolumn,letterpaper, xcolor=table]{article}
\usepackage{cvpr}              
\usepackage{siunitx}
\usepackage{multirow}
\usepackage{graphicx}
\usepackage{amsmath}
\usepackage{amssymb}
\usepackage{booktabs}

\usepackage{kotex}
\usepackage{verbatim}
\usepackage{amssymb}
\usepackage[table,xcdraw]{xcolor}

\usepackage[pagebackref,breaklinks,colorlinks]{hyperref}

\usepackage[capitalize]{cleveref}
\crefname{section}{Sec.}{Secs.}
\Crefname{section}{Section}{Sections}
\Crefname{table}{Table}{Tables}
\crefname{table}{Tab.}{Tabs.}

\def\cvprPaperID{*****} 
\def\confName{CVPR}
\def\confYear{2022}

\begin{document}

\title{Diffusion Probabilistic Models for Scene-Scale 3D Categorical Data}
\author{Jumin Lee\,\,\,\,\,\,
Woobin Im\,\,\,\,\,\,
Sebin Lee\,\,\,\,\,\,
Sung-Eui Yoon\\\\

Korea Advanced Institute of Science and Technology (KAIST)\\
{\small
\texttt{\{jmlee,iwbn,seb.lee,sungeui\}@kaist.ac.kr}}
}\maketitle

\begin{abstract}
   In this paper, we learn a diffusion model to generate 3D data on a scene-scale. Specifically, our model crafts a 3D scene consisting of multiple objects, while recent diffusion research has focused on a single object. 
To realize our goal, we represent a scene with discrete class labels, i.e., categorical distribution, to assign multiple objects into semantic categories. 
Thus, we extend discrete diffusion models to learn scene-scale categorical distributions. 
In addition, we validate that a latent diffusion model can reduce computation costs for training and deploying. 
To the best of our knowledge, our work is the first to apply discrete and latent diffusion for 3D categorical data on a scene-scale.
We further propose to perform semantic scene completion (SSC) by learning a conditional distribution using our diffusion model, where the condition is a partial observation in a sparse point cloud.
In experiments, we empirically show that our diffusion models not only generate reasonable scenes, but also perform the scene completion task better than a discriminative model.
Our code and models are available at \url{https://github.com/zoomin-lee/scene-scale-diffusion}. 
\end{abstract}

\section{Introduction}
\begin{figure}[t]
 \centering
\begin{subfigure}[b]{0.48\textwidth}
 \centering
 \includegraphics[trim={0cm 13.77cm 10.73cm 0cm},clip,width=1.0\columnwidth]{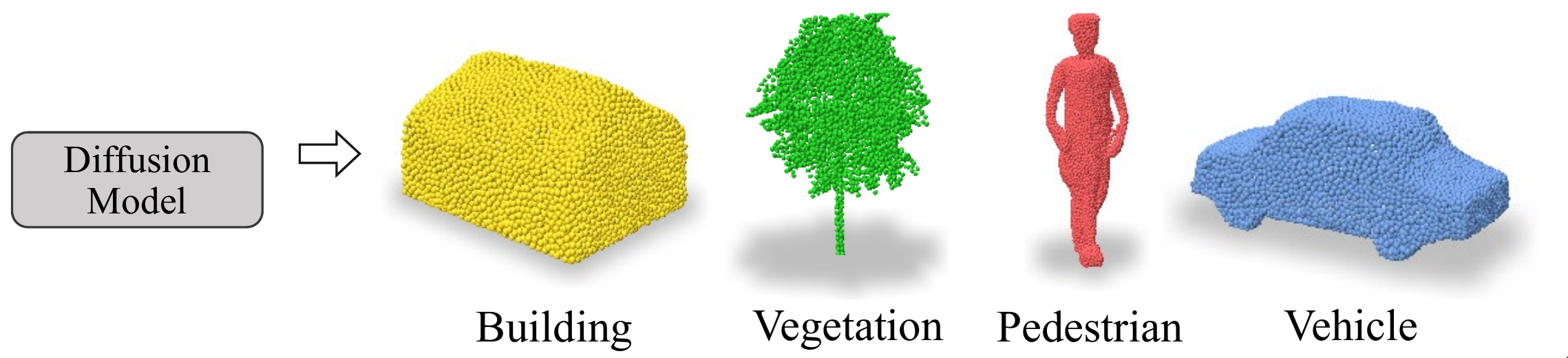}
 \caption{Object-scale generation}\label{subfig:pointe}
 \end{subfigure}
 \hfill
\begin{subfigure}[b]{0.48\textwidth}
 \centering
 \includegraphics[trim={0cm 10.04cm 10.73cm 0cm},clip,width=1.0\columnwidth]{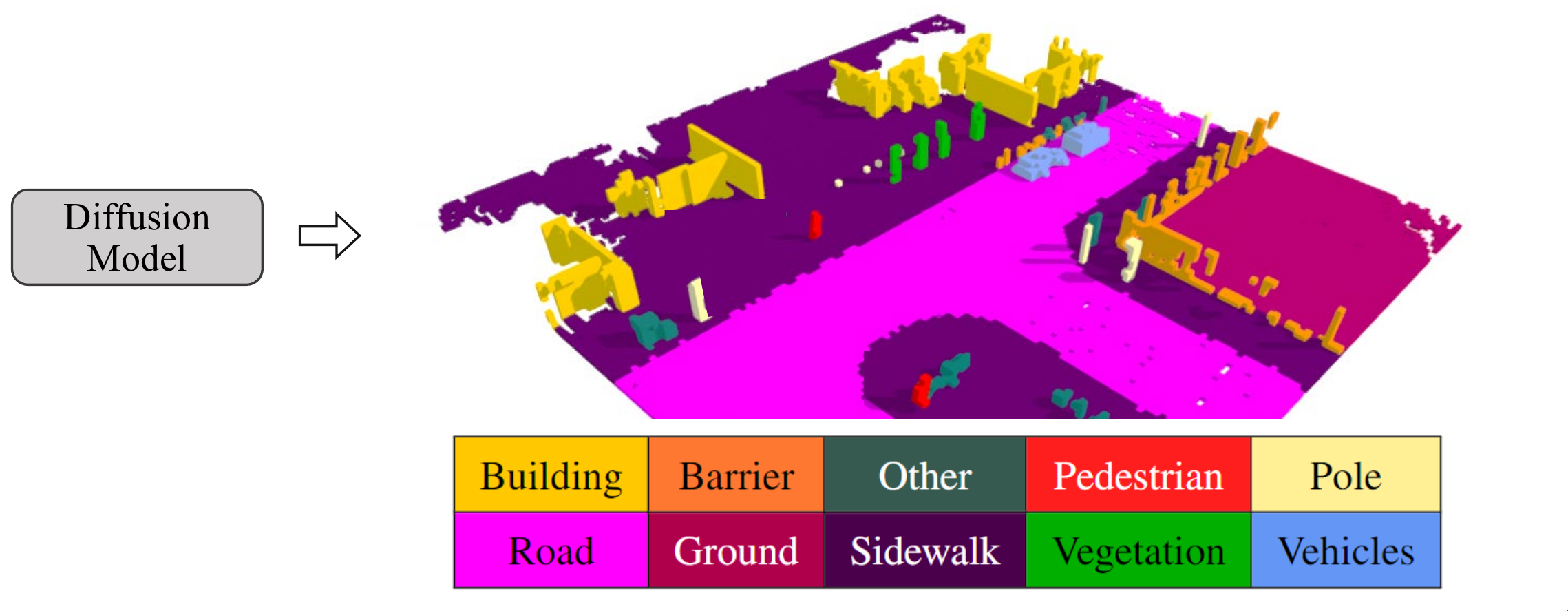}
 \caption{Scene-scale generation (\textbf{ours})}\label{subfig:scene}
\end{subfigure}
\centering
\caption{Comparison of object-scale and scene scale generation (ours). Our result includes multiple objects in a generated scene, while the object-scale generation crafts one object at a time. (a) is obtained by Point-E~\cite{pointe}.
}
\label{fig:1}
\end{figure}

Learning to generate 3D data has received much attention thanks to its high performance and promising downstream tasks. 
For instance, a 3D generative model with a diffusion probabilistic model~\cite{c3d} has shown its effectiveness in 3D completion~\cite{c3d} and text-to-3D generation~\cite{lion,pointe}. 

While recent models have focused on 3D object generation, 
we aim beyond a single object by generating a 3D scene with multiple objects. 
In Fig.~\ref{subfig:scene}, we show a sample scene from our generative model, 
where we observe the plausible placement of the objects, 
as well as their correct shapes. 
Compared to the existing object-scale model~\cite{pointe} (Fig.~\ref{subfig:pointe}), 
our scene-scale model can be used in a broader application, 
such as semantic scene completion (Sec.~\ref{sec:expr_ssc}), 
where we complete a scene given a sparse LiDAR point cloud.

We base our scene-scale 3D generation method on a diffusion model, which has shown remarkable performance in modeling complex real-world data, such as realistic 2D images~\cite{firstddp,ddpm,beat} and 3D objects~\cite{c3d,lion,pointe}.
We develop and evaluate diffusion models learning a scene-scale 3D categorical distribution.

First, we utilize categorical data for a voxel entity since we have multiple objects in contrast to the existing work~\cite{c3d,lion,pointe}, so each category tells each voxel belongs to which category.
Thus, we extend discrete diffusion models for 2D categorical data~\cite{multinomal_diffusion,discrete_diffusion} 
into  3D categorical data (Sec.~\ref{method_discrete}).
Second, we validate the latent diffusion model for the 3D scene-scale generation, which can reduce training and testing computational cost (Sec.~\ref{method_latent}).
Third, we propose to perform semantic scene completion (SSC) by 
learning a conditional distribution using our generative models,
where the condition is a partial observation of the scene (Sec.~\ref{method_discrete}).
That is, we demonstrate that our model can complete a reasonable scene in
a realistic scenario with a sparse and partial observation.

Lastly, we show the effectiveness of our method in terms of the unconditional and conditional (SSC) generation tasks on the CarlaSC dataset~\cite{carlasc} (Sec.~\ref{sec:expr}). Especially, we show that our generative model can outperform a discriminative model in the SSC task.

\section{Related Work}
\subsection{Semantic Scene Completion}
\label{sec:twoCond}

Leveraging 3D data for semantic segmentation has been studied from different perspectives.
Vision sensors (e.g., RGB-D camera and LiDAR) provide
depth information from a single viewpoint, giving more information about the world. One of the early approaches is using an RGB-D (i.e., color and depth) image with a 2D segmentation map~\cite{ss}.
In addition, using data in a 3D coordinate system has been extensively studied.
3D semantic segmentation is the extension of 2D segmentation, where a classifier is applied to point clouds or voxel data in 3D coordinates~\cite{qi2017pointnet,riegler2017octnet}.

One of the recent advances in 3D semantic segmentation is semantic scene completion (SSC), where
a partially observable space -- observed via RGB-D image or point clouds -- should be densely filled with class labels~\cite{s3cnet,js3c,implicit,lmsc}.
In SSC, a model gets the point cloud obtained in one viewpoint; thus, it contains multiple partial objects (e.g., one side of a car). Then, the model not only reconstructs the unobserved shape of the car but also labels it as a car.
Here, the prediction about the occupancy and the semantic labels can mutually benefit~\cite{SSC}.

Due to the partial observation, filling in occluded and sparse areas is the biggest hurdle.
Thus, a generative model is effective for 3D scene completion as 2D completion tasks~\cite{jo2021n,lugmayr2022repaint}.
Chen et al.~\cite{scenegan} demonstrate that generative adversarial networks~(GANs) can be used to improve the plausibility of a completion result.
However, a diffusion-based generative model has yet to be explored in terms of a 3D semantic segmentation map. 
We speculate that using a diffusion model has good prospects, thanks to the larger size of the latent and the capability to deal with high-dimensional data.

In this work, we explore a diffusion model in the context of 3D semantic scene completion.
Diffusion models have been rapidly growing and they perform remarkably well on real-world 2D images~\cite{ramesh2022hierarchical}.
Thus, we would like to delve into the diffusion to generate 3D semantic segmentation maps; thus, we hope to provide the research community a useful road map towards generating the 3D semantic scene maps.

\subsection{Diffusion Models}
\label{sec:diffusion}

Recent advances in diffusion models
have shown that a deep model can learn more diverse data distribution by a diffusion process~\cite{ddpm}.
A diffusion process is introduced to adopt a simple distribution (e.g., Gaussian) to learn a complex distribution~\cite{firstddp}.
Especially, diffusion models show impressive results for image generation~\cite{beat} and conditional generation~\cite{palette, text} on high resolution compared to GANs. 
GANs are known to suffer from the mode collapse problem and struggle to capture complex scenes with multiple objects~\cite{sceneimagegan}.
On the other hand, diffusion models have a capacity to escape mode collapse~\cite{beat} and generate complex scenes~\cite{text, frido} since likelihood-based methods achieve better coverage of full data distribution.

Diffusion models have been studied to a large extent in high-dimensional continuous data.
However, they often lack the capacity to deal with discrete data (\textit{e.g.}, text and segmentation maps) 
since the discreteness of data is not fully covered by continuous representations.
To tackle such discreteness, discrete diffusion models have been studied for various applications, 
such as text generation~\cite{discrete_diffusion, multinomal_diffusion} and low-dimensional segmentation maps generation~\cite{multinomal_diffusion}.

Since both continuous and discrete diffusion models estimate the density of image pixels, a higher image resolution means higher computation. 
To address this issue, latent diffusion models~\cite{stable, text} operate a diffusion process on the latent space of a lower dimension.
To work on the compressed latent space, Vector-Quantized Variational Auto-Encoder (VQ-VAE)~\cite{vqvae} is employed. 
Latent diffusion models consist of two stages: VQ-VAE and diffusion.
VQ-VAE trains an encoder to compress the image into a latent space.
Equipped with VQ-VAE, autoregressive models~\cite{vqgan,dalle} have shown impressive performance.
Recent advances in latent diffusion models further improve the generative performance by ameliorating the unidirectional bias and accumulated prediction error in existing models~\cite{text, stable}.

Our work introduces an extension of discrete diffusion models for high-resolution 3D categorical voxel data. Specifically, we show the effectiveness of a diffusion model in terms of unconditional and conditional generation tasks, where the condition is a partial observation of a scene (\textit{i.e.}, SSC). Further, we propose a latent diffusion models for 3D categorical data to reduce the computation load caused by high-resolution segmentation maps.

\subsection{Diffusion Models for 3D Data}
\label{sec:3ddiffusion}
Diffusion models have been used for 3D data.
Until recently, research has been mainly conducted for 3D point clouds with \textit{xyz}-coordinates.
PVD~\cite{c3d} applies continuous diffusion on point-voxel representations for object shape generation and completion without additional shape encoders.
LION~\cite{lion} uses latent diffusion for object shape completion (\textit{i.e.}, conditional generation) with additional shape encoders.

In this paper, we aim to learn 3D categorical data (\textit{i.e.}, 3D semantic segmentation maps) with a diffusion model. 
The study of object generation has shown promising results, but as far as we know, our work is the first to generate a 3D scene with multiple objects using a diffusion model.
Concretely, our work explores discrete and latent diffusion models to learn a distribution of volumetric semantic scene segmentation maps. 
We develop the models in an unconditional and conditional generation; the latter can be used directly for the SSC task.

\section{Method}
\begin{figure}[t]
 \centering
\begin{subfigure}[b]{0.48\textwidth}
 \centering
 \includegraphics[trim={0cm 6.58cm 0cm 0cm},clip,width=1.0\columnwidth]{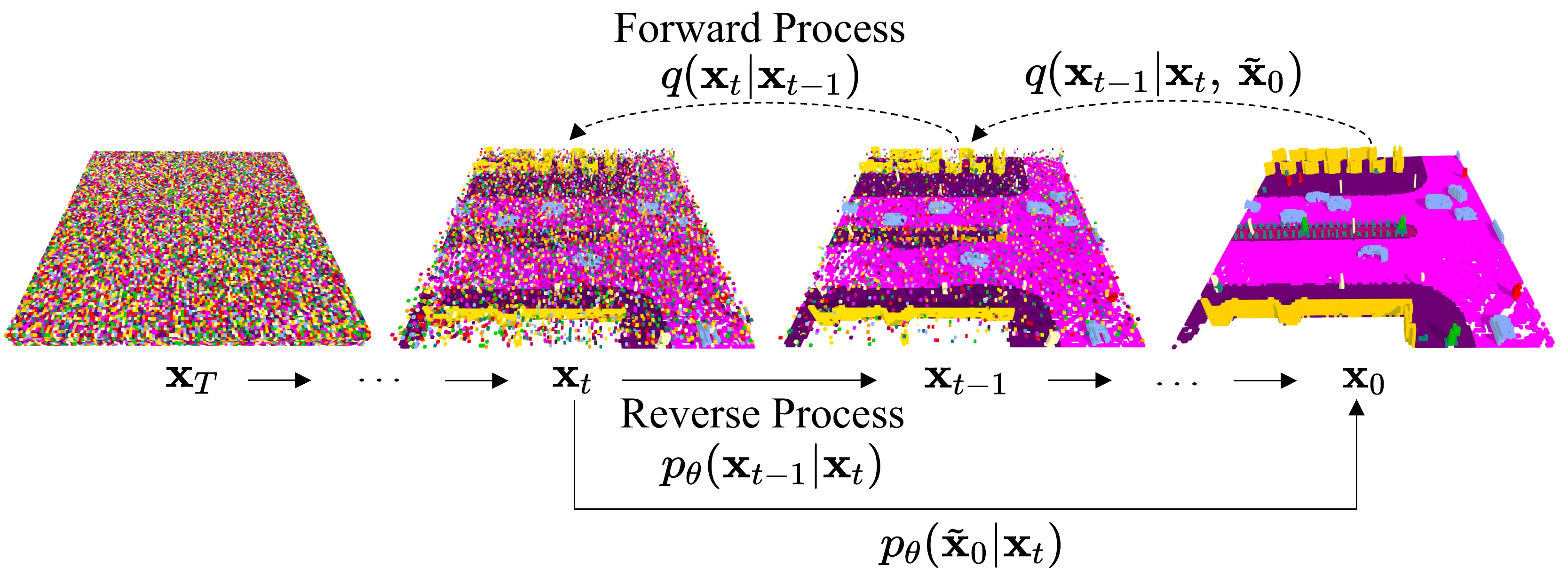}
 \caption{Discrete Diffusion Models}\label{subfig:discrete}
 \end{subfigure}
 \hfill
\begin{subfigure}[b]{0.48\textwidth}
 \centering
 \includegraphics[trim={0cm 0cm 6.9cm 0cm},clip,width=1.0\columnwidth]{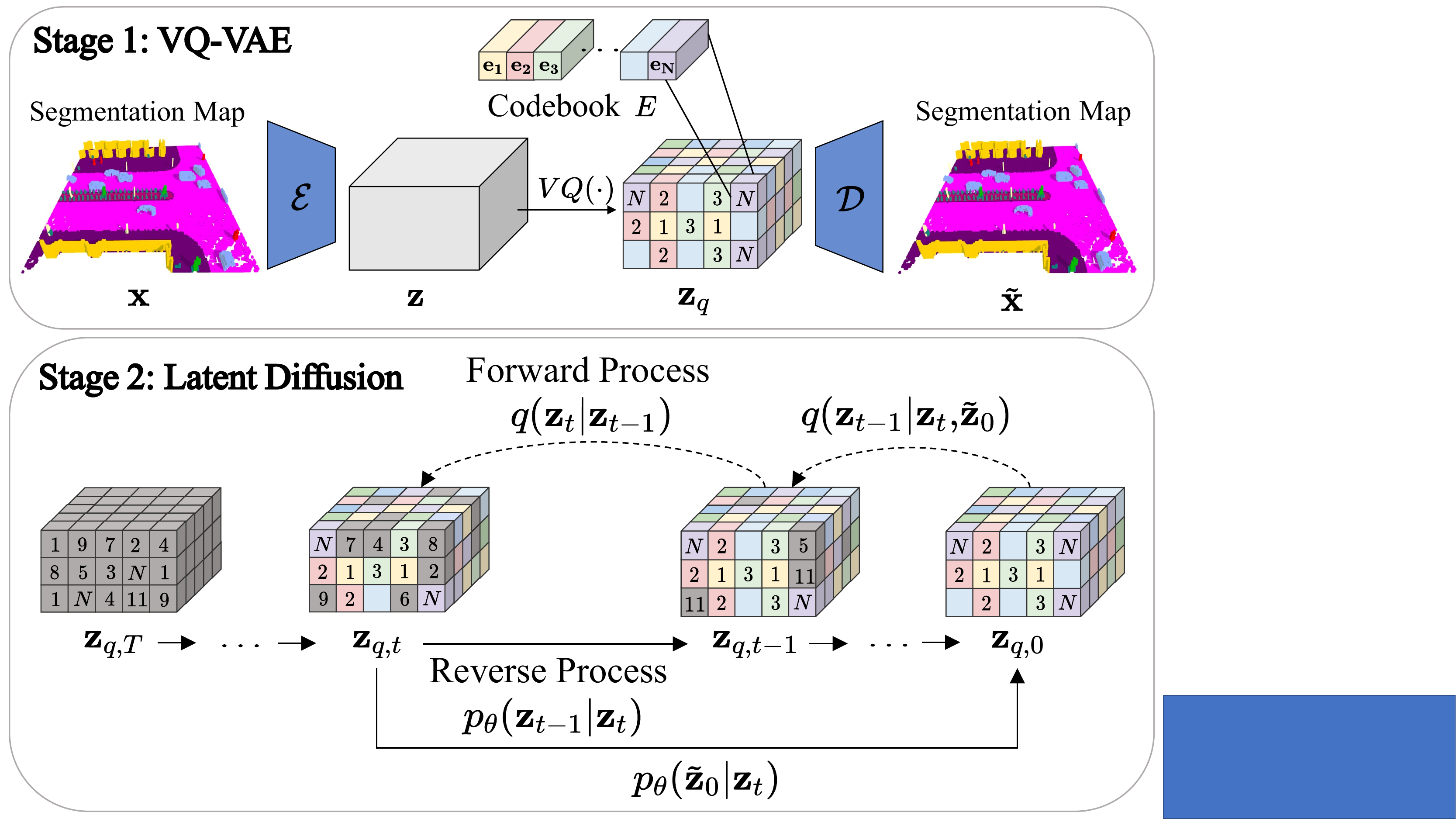}
 \caption{Latent Diffusion Models}\label{subfig:latent}
\end{subfigure}
\centering
\caption{Overview of (a) Discrete Diffusion Models and (b) Latent Diffusion Models.
Discrete diffusion models conduct diffusion process on voxel space, whereas latent diffusion models operate diffusion process on latent space. 
}
\label{fig:1}
\end{figure}

Our goal is to learn a data distribution $p(\mathbf{x})$ using diffusion models, where each data $\mathbf{x}\sim p(\mathbf{x})$ represents a 3D segmentation map described with the one-hot representation. 
3D segmentation maps are  samples from the data distribution $p(\mathbf{x}$), which is the categorical distribution $\text{Cat}(k_0, k_1, \cdots, k_M)$ with $M+1$ probabilities of the free label $k_0$ and $M$ main categories.
The discrete diffusion models could learn data distribution by recovering the noised data, which is destroyed through the successive transition of the label~\cite{discrete_diffusion}.

Our method aims to learn a distribution of voxelized 3D segmentation maps with discrete diffusion (Sec.~\ref{method_discrete}).
Specifically, it includes
unconditional and conditional generation, where the latter corresponds to the SSC task.
In addition, we explore a latent diffusion model for 3D segmentation maps (Sec.~\ref{method_latent}).

\subsection{Discrete Diffusion Models}
\label{method_discrete}
Fig.~\ref{subfig:discrete} summarizes the overall process of discrete diffusion, consisting of a forward process and a reverse process; the former gradually adds noise to the data and the latter learns to denoise the noised data. 

In the forward process in the discrete diffusion, 
an original segmentation map $\mathbf{x}_0$ is gradually corrupted into a $t$-step noised segmentation map $\mathbf{x}_t$ with $1 \le t \le T$.
Each forward step can be defined by a Markov uniform transition matrix $Q_{t}$~\cite{discrete_diffusion} as $\mathbf{x}_t = \mathbf{x}_{t-1} Q_t$.
Based on the Markov property, we can derive the $t$-step noised segmentation map $\mathbf{x}_{t}$ straight from the original segmentation map $\mathbf{x}_{0}$, $q(\mathbf{x}_{t}|\mathbf{x}_{0})$, with a cumulative transition matrix $\bar{Q}_{t}=Q_{1}Q_{2}\cdots Q_{t}$: 
\begin{equation}
\label{eq:1}
q(\mathbf{x}_{t}|\mathbf{x}_{0}) = \text{Cat}(\mathbf{x}_{t} ; p=\mathbf{x}_{0}\bar{Q}_{t}).
\end{equation}

In the reverse process parametrized by $\theta$, a learnable model is used to reverse a noised segmentation map by $p_{\theta}(\mathbf{x}_{t-1}|\mathbf{x}_{t})$.
Specifically, we use a reparametrization trick~\cite{ddpm} to make the model predict a denoised map $\tilde{\mathbf{x}}_{0}$ and subsequently get the reverse process $p_{\theta}(\mathbf{x}_{t-1}|\mathbf{x}_{t})$:
\begin{align}
p_{\theta}(\mathbf{x}_{t-1}|\mathbf{x}_{t}) &= q(\mathbf{x}_{t-1}| \mathbf{x}_{t},\tilde{\mathbf{x}}_{0}) p_{\theta}(\tilde{\mathbf{x}}_{0}|\mathbf{x}_{t}),
\label{eq:2}\\
q(\mathbf{x}_{t-1}|\mathbf{x}_{t},\tilde{\mathbf{x}}_{0}) &= \frac{q(\mathbf{x}_{t}| \mathbf{x}_{t-1},\tilde{\mathbf{x}}_{0}) q(\mathbf{x}_{t-1}|\tilde{\mathbf{x}}_{0})}{q(\mathbf{x}_{t}| \tilde{\mathbf{x}}_{0})}.\label{eq:3}
\end{align}

We optimize a joint loss that consists of the KL divergence of the forward process $q(\mathbf{x}_{t-1}|\mathbf{x}_{t},\mathbf{x}_{0})$ from the reverse process $p_\theta(\mathbf{x}_{t-1}|\mathbf{x}_t)$; of the original segmentation map $q(\mathbf{x}_{0})$ from the reconstructed one $p_\theta(\mathbf{x}_{t-1}|\mathbf{x}_t)$ for an auxiliary loss:
\begin{multline}
\label{eq:loss}
\mathcal{L} = D_{KL}(\, q(\mathbf{x}_{t-1}|\mathbf{x}_t, \mathbf{x}_0) \,\| \, p_\theta(\mathbf{x}_{t-1}|\mathbf{x}_t)\,)
\\ + w_0 D_{KL}(\, q(\mathbf{x}_{0}) \,\| \, p_{\theta}(\tilde{\mathbf{x}}_{0}|\mathbf{x}_{t}) \,),
\end{multline}
where $w_0$ is an auxiliary loss weight.

Unlike existing discrete diffusion models~\cite{multinomal_diffusion,discrete_diffusion},
our goal is to learn the distribution of 3D data. Thus, to better handle 3D data, we use a point cloud segmentation network~\cite{cylinder} with modifications for discrete data and time embedding.
\paragraph{Conditional generation.} 
We propose discrete diffusion for Semantic Scene Completion~(SSC) with conditional generation.
SSC jointly estimates a scene's complete geometry and semantics, given a sparse occupancy map $\mathbf{s}$.
Thus, it introduces a condition into Eq.~\ref{eq:2}, resulting in:
\begin{equation}
\label{eq:4}
p_{\theta}(\mathbf{x}_{t-1}|\mathbf{x}_{t}, \mathbf{s}) = q(\mathbf{x}_{t-1}| \mathbf{x}_{t},\tilde{\mathbf{x}}_{0}) p_{\theta}(\tilde{\mathbf{x}}_{0}|\mathbf{x}_{t}, \mathbf{s}),
\end{equation}
where $\mathbf{s}$ is a sparse occupancy map. 
We give the condition by concatenating a sparse occupancy map $\mathbf{s}$ with a corrupted input $\mathbf{x}_t$.

\subsection{Latent Diffusion Models}
\label{method_latent}
Fig.~\ref{subfig:latent} provides an overview of latent diffusion on 3D segmentation maps.
Latent diffusion models project the 3D segmentation maps into a smaller latent space and operate a diffusion process on the latent space instead of the high-dimensional input space. A latent diffusion takes advantage of a lower training computational cost and a faster inference by processing diffusion on a lower dimensional space.

To encode a 3D segmentation map into a latent representation, we use Vector Quantized Variational AutoEncoder~(VQ-VAE)~\cite{vqvae}. 
VQ-VAE extends the VAE by adding a discrete learnable codebook $E=\{\mathbf{e}_n\}^{N}_{n=1} \in \mathbb{R}^{N\times d}$, where $N$ is the size of the codebook and $d$ is the dimension of the codes.
The encoder $\mathcal{E}$ encodes 3D segmentation maps $\mathbf{x}$ into a latent $\mathbf{z}=\mathcal{E}(\mathbf{x})$, and the quantizer $VQ(\cdot)$ maps the latent $\mathbf{z}$ into a quantized latent $\mathbf{z}_q$, which is the closest codebook entry $\mathbf{e}_n$. 
Note that the latent $\mathbf{z}\in \mathbb{R}^{h\times w\times z\times d}$ has a smaller spatial resolution than the segmentation map $\mathbf{x}$.
Then the decoder $\mathcal{D}$ reconstructs the 3D segmentation maps from the quantized latent, $\tilde{\mathbf{x}} = \mathcal{D}(VQ(\mathcal{E}(\mathbf{x})))$. The encoder $\mathcal{E}$, the decoder $\mathcal{D}$, and
the codebook $E$ can be trained end-to-end using the following loss function:
\begin{multline}
\label{eq:vqvae_loss}
\mathcal{L}_{VQVAE} = - \sum_k w_k \mathbf{x}_k  \log (\tilde{\mathbf{x}}_k) + \|sg(\mathbf{z})-\mathbf{z}_q\|_2^2 \\ + \|\mathbf{z}-sg(\mathbf{z}_q)\|_2^2,
\end{multline} 
where $w_k$ is a class weight and $sg(\cdot)$ is the stop-gradient operation.
Training the latent diffusion model is similar to that of discrete diffusion.
Discrete diffusion models diffuse between labels, but latent diffusion models diffuse between codebook indexes using Markov Uniform transition matrix $Q_{t}$~\cite{discrete_diffusion}. 

\section{Experiments}\label{sec:expr}

In this section, we empirically study the effectiveness of the diffusion models on 3D voxel segmentation maps.
We divide the following sub-sections into the learning of the unconditional data distribution $p(\mathbf{x})$ (Sec.~\ref{sec:expr_3dseg}) and the conditional data distribution $p(\mathbf{x}|\mathbf{s})$ 
given a sparse occupancy map $\mathbf{s}$ (Sec.~\ref{sec:expr_ssc}); note that the latter corresponds to semantic scene completion (SSC).

\subsection{Implementation Details}

\noindent \textbf{Dataset.} Following prior work~\cite{carlasc}, we employ the CarlaSC dataset -- a synthetic outdoor driving dataset -- for training and evaluation.
The dataset consists of 24 scenes in 8 dynamic maps under low, medium, and high traffic conditions. 
The splits of the dataset contain 18 training, 3 validation, and 3 test scenes, which are annotated with 10 semantic classes and a free label.
Each scene with a resolution of $128\times128\times8$ covers a range of \SI{25.6}{\meter} ahead and behind the car, \SI{25.6}{\meter} to each side, and \SI{3}{\meter} in height.

\noindent \textbf{Metrics.}
Since SSC requires predicting the semantic label of a voxel and an occupancy state together, 
we use mIoU and IoU as SSC and VQ-VAE metrics. The mIoU measures the intersection over union averaged 
over all classes, and the IoU evaluates scene completion quality, 
regardless of the predicted semantic labels.

\noindent \textbf{Experimental settings.}
Experiments are deployed on two NVIDIA GTX 3090 GPUs with a batch size of 8 for diffusion models and 4 for VQ-VAE.
Our models follow the same training strategy as multinomial diffusion~\cite{multinomal_diffusion}.
We set the hyper-parameters of the diffusion models with the number of time steps $T=100$ timesteps. And for VQ-VAE, we set the codebook $E = \{\mathbf{e}_n\}^{N}_{n=1} \in \mathbb{R}^{N\times d}$ where the codebook size $N=1100$, dimension of codes $d=11$ and $\mathbf{e}_n\in \mathbb{R}^{32\times32\times2\times d}$.
For diffusion architecture, we slightly modify the encoder--decoder structure in Cylinder3D~\cite{cylinder} for time embedding and discreteness of the data. And for VQ-VAE architecture, we also use encoder--decoder structure in Cylinder3D~\cite{cylinder}, but with the vector quantizer module.

\subsection{3D Segmentation Maps Generation}\label{sec:expr_3dseg}
\begin{figure*}[p]
 \centering
\centerline{\includegraphics[trim={0cm 10.3cm 0cm 0cm},clip,width=1.0\linewidth]{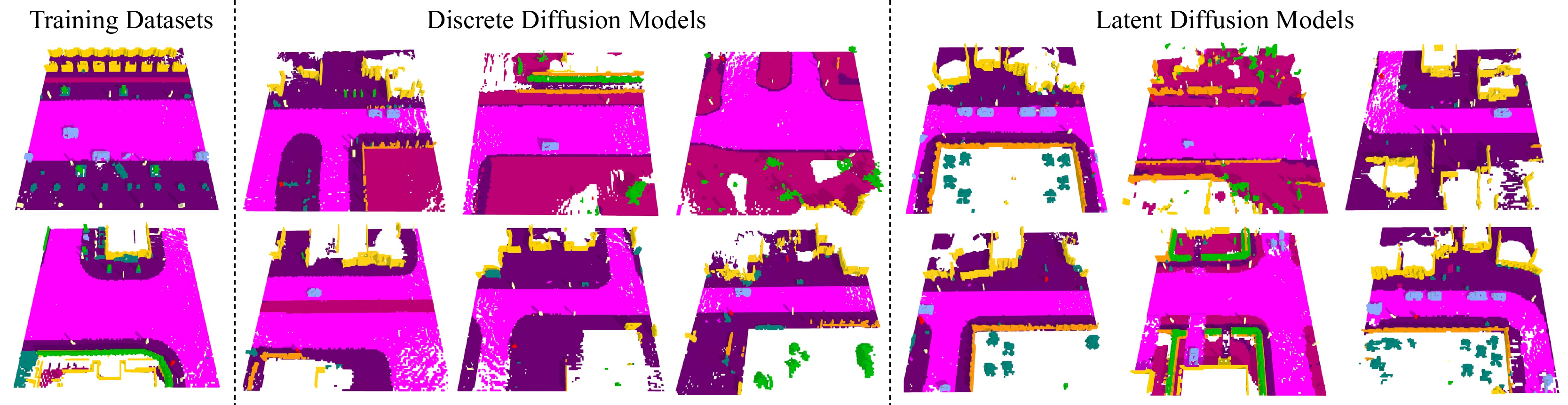}}
\caption{
    \textbf{Samples from our unconditional diffusion models.} The first column shows samples from training datasets. From the second column, we show samples from our discrete diffusion and latent diffusion models. 
    We can observe our diffusion models learn the 3D categorical distribution well, so that it is capable to generate a variety of plausible maps. Color assignment for each class is available in Tab.~\ref{tab:ours}. 
}\label{fig:generation}
\end{figure*}

We use the discrete and the latent diffusion models for 3D segmentation map generation.
Fig.~\ref{fig:generation} shows the qualitative results of the generation.
As seen in the figure, both the discrete and latent models learn the categorical distribution as they produce a variety of reasonable scenes. 
Note that our models are learned on a large-scale data distribution like the 3D scene with multiple objects; this is worth noting since recent 3D diffusion models for point clouds have been performed on an object scale~\cite{c3d, m3d, l3d, lion}.

\begin{table}[t]
\renewcommand{\arraystretch}{1.1}
\small \centering
\begin{tabular}{c|c|c|c}
\hline
Model & Resolution & \begin{tabular}[c]{@{}c@{}}Training\\ (time/epoch)\end{tabular} & \begin{tabular}[c]{@{}c@{}}Sampling\\ (time/img)\end{tabular} \\ \hline
D-Diffusion & 128$\times$128$\times$8 & 19m 48s & 0.883s \\ \hline
\multirow{3}{*}{L-Diffusion} & 32$\times$32$\times$2 & 7m 37s & 0.499s \\
                        & 16$\times$16$\times$2 & 4m 41s & 0.230s \\
                        & 8$\times$8$\times$2   & 4m 40s &  0.202s \\ \hline
\end{tabular}
\caption{\textbf{Computation time comparison} between discrete diffusion models and latent diffusion models for 3D segmentation maps generation. `D-Diffusion' and `L-Diffusion' denote discrete diffusion models and latent diffusion models, respectively. `Resolution' means the resolution of the space in which diffusion process operates. A latent diffusion models process diffusion on a lower dimensional latent space, as a result, it shows
advantage of a faster training and sampling time.}
\label{tab:time}
\end{table}
In Tab.~\ref{tab:time}, we compare training and sampling time models for different resolutions on which each diffusion model operates. 
Compared to the discrete diffusion, the latent diffusion tends to show shorter training and inference time.
This is because the latent diffusion models compress the data into a smaller latent so that the time decreases as the compression rate increases.
In particular, compared to discrete diffusion, which performs a diffusion process in voxel space, $32\times32\times32$ latent diffusion has 2.6~times faster training time for one epoch and 1.8~times faster sampling time for generating one image.

\begin{table}[t]
\small \centering
\begin{tabular}{c|c|c|c}
\hline
\begin{tabular}[c]{@{}c@{}}Codebook size\\ ($N$)\end{tabular} & \begin{tabular}[c]{@{}c@{}}Resolution\\ ($h \times w \times z$)\end{tabular} & IoU & mIoU \\ \hline
\multirow{3}{*}{220} & 8$\times$8$\times$2       & \multicolumn{1}{c|}{72.5} & 27.3 \\
                      & 16$\times$16$\times$2     & \multicolumn{1}{c|}{78.7} & 36.9  \\
                      & 32$\times$32$\times$2     & \multicolumn{1}{c|}{84.6} & 56.5 \\ \hline
\multirow{3}{*}{550} & 8$\times$8$\times$2       & \multicolumn{1}{c|}{67.7} & 25.7 \\
                      & 16$\times$16$\times$2     & \multicolumn{1}{c|}{79.4} & 39.7   \\
                      & 32$\times$32$\times$2     & \multicolumn{1}{c|}{85.8} & 58.4 \\ \hline
\multirow{3}{*}{1,100}& 8$\times$8$\times$2       & \multicolumn{1}{c|}{70.3} & 25.7 \\
                     & 16$\times$16$\times$2     & \multicolumn{1}{c|}{79.3} & 35.0 \\
                      & 32$\times$32$\times$2     & \multicolumn{1}{c|}{89.1} & 65.1 \\ \hline
\multirow{3}{*}{2,200}& 8$\times$8$\times$2       & \multicolumn{1}{c|}{70.2} & 26.5 \\
                      & 16$\times$16$\times$2     & \multicolumn{1}{c|}{77.7} & 37.9 \\
                      & 32$\times$32$\times$2    & \multicolumn{1}{c|}{89.2} & 64.2 \\ \hline
\end{tabular}

\caption{\textbf{Ablation study on VQ-VAE hyper-parameters.} We compare different sizes of codebook $N$ and resolutions of the latent space $h{\times}w{\times}z$.}
\label{tab:ablation}
\end{table}
\paragraph{Ablation study on VQ-VAE.} 
Latent diffusion models consist of two stages. 
The VQ-VAE compresses 3D segmentation maps to latent space, and then discrete diffusion models apply on the codebook index of latent. 
Therefore, the performance of VQ-VAE may set the upper bound for the final generation quality. 
So we conduct an ablation study about VQ-VAE while adjusting the resolution of the latent space $h\times w\times z$ and the codebook capacities $N$ while keeping the code dimension $d$ fixed.
Concretely, we compress the 3D segmentation maps from $128\times128\times8$ to $32\times32\times2$, $16\times16\times2$, and $8\times8\times2$ with four different codebook size $N \in \{220, 550, 1100, 2200\}$.

The quantitative comparison is shown in Tab.~\ref{tab:ablation}.
The bigger the codebook size is, the higher the performance is, but it saturates  around 1,100. 
That is because most of the codes are not updated, and the update of the codebook can lapse into a local optimum~\cite{global}. 

The resolution of latent space has a significant impact on performance. As the resolution of the latent space becomes smaller, it cannot contain all the information of the 3D segmentation map.
Setting the resolution to $32\times32\times2$ with a $1{,}100$ codebook size strike a good balance between efficiency and fidelity.

\begin{table}[t]
\small \centering
\centering
\begin{tabular}{c|c|c}
\hline
Methods & IoU & mIoU \\ \hline
LMSCNet SS~\cite{lmsc} & 85.98 & 42.53 \\
SSCNet Full~\cite{SSC} & 80.69 & 41.91 \\
MotionSC (T=1)~\cite{carlasc} & 86.46 & 46.31 \\ \hline
Our network w/o Diffusion & 80.70 & 39.94 \\
Discrete Diffusion (Ours) & 80.61 & 45.83 \\ \hline
\end{tabular}
\caption{Semantic Scene Completion results on test set of CarlaSC}
\label{tab:compare}
\end{table}

\begin{table*}[p]{
\centering
\renewcommand{\arraystretch}{1.1}
\resizebox{\textwidth}{!}
{\normalsize
\begin{tabular}{c|cccccccccccc|c}
\hline
 & \multicolumn{12}{c|}{Class IoU} &  \\ \cline{2-13}
\multirow{-2}{*}{} & \multicolumn{1}{c|}{mIoU} & Free & \cellcolor[HTML]{FFC800}Building & \cellcolor[HTML]{FF7832}Barrier & \cellcolor[HTML]{375A50}{\color[HTML]{FFFFFF} Other} & \cellcolor[HTML]{FF1E1E}{\color[HTML]{FFFFFF}Pedestrian} & \cellcolor[HTML]{FFF096}Pole & \cellcolor[HTML]{FF00FF}Road & \cellcolor[HTML]{AF004B}{\color[HTML]{FFFFFF} Ground} & \cellcolor[HTML]{4B004B}{\color[HTML]{FFFFFF} Sidewalk} & \cellcolor[HTML]{00AF00}Vegetation & \cellcolor[HTML]{6496F5}Vehicles & \multirow{-2}{*}{IoU} \\ \hline
w/o Diffusion & \multicolumn{1}{c|}{39.94} & 96.40 & 27.72 & 3.15 & 8.77 & 22.15 & 37.14 & \textbf{89.02} & \textbf{18.22} & 59.25 & 29.74 & 47.72 & 80.70 \\
Discrete Diffusion (Ours) & \multicolumn{1}{c|}{\textbf{45.83}} & 96.00 & \textbf{31.75} & 3.42 & \textbf{25.43} & \textbf{46.22} & \textbf{43.32} & 84.57 & 13.01 & \textbf{67.50} & \textbf{37.45} & \textbf{55.46} & 80.61 \\ \hline
\end{tabular}
}
}
\caption{\textbf{Semantic scene completion results} on test set of CarlaSC. The discriminative learning result with the diffusion model architecture is denoted as `w/o Diffusion'. 
Values with a difference equal to or greater than 0.5\%p are bold.}
\label{tab:ours}
\end{table*}
\begin{figure*}[p]
\begin{center}
\centerline{\includegraphics[trim={0cm 0cm 0cm 0cm},clip,width=1.0\linewidth]{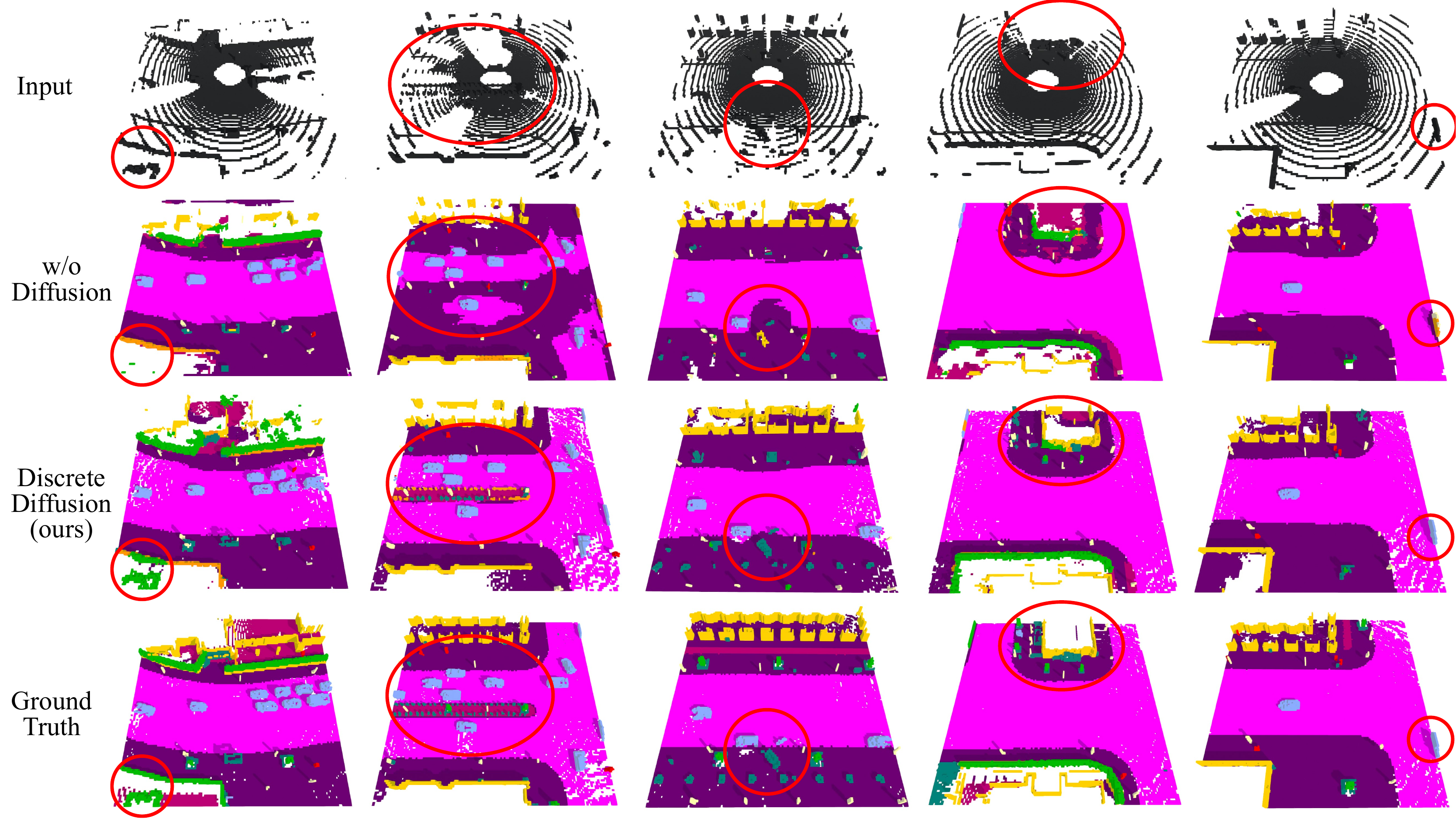}}
\caption{\textbf{Qualitative comparison of a deterministic model (w/o diffusion) and ours (discrete diffusion) on the test split of CarlaSC.}
The first row shows the sparse inputs for the scene completion task, and the last row shows the corresponding ground-truth. Compared to the deterministic model, our probabilistic model produces more plausible shape and class inference, as highlighted by the red circles. Note that the both models (w/o diffusion and discrete diffusion) use the same network architecture. Color assignment for each class is available in Tab.~\ref{tab:ours}. }
\label{fig:completion}
\end{center}
\end{figure*}
\subsection{Semantic Scene Completion}\label{sec:expr_ssc}

We use a discrete diffusion model for conditional 3D segmentation map generation (\textit{i.e.}, SSC).
As a baseline model against the diffusion model, we train a network with an identical architecture by discriminative learning without a diffusion process.
We optimize the baseline with a loss term $\mathcal{L} = - \sum_k w_k \mathbf{x}_k  \log (\tilde{\mathbf{x}}_k)$, where $w_k$ is a weight for each semantic class.
We visualize results from the baseline and our discrete diffusion model in Fig.~\ref{fig:completion}.
Despite the complexities of the networks being identical, our discrete diffusion model improves mIoU (\textit{i.e.}, class-wise IoU) up to 5.89\%p than the baseline model as shown in Tab.~\ref{tab:ours}. 
Especially, our method achieves outstanding results in small objects and fewer frequency categories like `pedestrian', `pole', `vehicles,' and `other'.
The qualitative results in Fig.~\ref{fig:completion} better demonstrate the improvement. 

In Tab.~\ref{tab:compare}, we compare our model with existing SSC models whose network architectures and training strategies are specifically built for the SSC task. 
Nonetheless, our diffusion model outperforms LMSCNet~\cite{lmsc} and SSCNet~\cite{SSC}, in spite of the simpler architecture and training strategies. Although MotionSC~\cite{carlasc} shows a slightly better result, we speculate that the diffusion probabilistic model can be improved by extensive future research dedicated to this field.

\section{Conclusion}
In this work, we demonstrate the extension of the diffusion model to scene-scale 3D categorical data beyond generating a single object.
We empirically show that our models have impressive generative power to craft various scenes through a discrete and latent diffusion process.
Additionally, our method provides an alternative view for the SSC task, showing superior performance compared to a discriminative counterpart.
We believe that our work can be a useful road map for generating 3D data with a diffusion model.

\bibliographystyle{IEEEtran}
\bibliography{bib}

\end{document}